\documentclass[10pt,a4paper,conference]{IEEEtran}

\usepackage{times}
\usepackage{epsfig}
\usepackage{graphicx}
\usepackage{amsmath}
\usepackage{amssymb}
\usepackage{verbatim}
\usepackage{booktabs} 
\usepackage{adjustbox}
\usepackage{colortbl} 
\usepackage{xcolor} 
\usepackage{xfrac}
\usepackage[]{algorithm2e}
\usepackage{subcaption}
\usepackage{graphicx,multirow}
\usepackage{array}
\usepackage{colortbl}

\usepackage{booktabs,subcaption,amsfonts,dcolumn}
\newcolumntype{d}[1]{D..{#1}}
\ifCLASSINFOpdf
\else
\fi
\begin{document}
%
\title{Multi-Level Feature Abstraction from Convolutional Neural Networks for Multimodal Biometric Identification}

\author{Sobhan Soleymani,  Ali Dabouei, Hadi Kazemi, Jeremy Dawson, and Nasser M. Nasrabadi, {\it Fellow, IEEE}\\
West Virginia University\\
{\tt\small \{ssoleyma, ad0046, hakazemi\}@mix.wvu.edu, \{jeremy.dawson, nasser.nasrabadi\}@mail.wvu.edu}}


%


\maketitle

\begin{abstract}
In this paper, we propose a deep multimodal fusion network to fuse multiple modalities (face, iris, and fingerprint) for person identification. The proposed deep multimodal fusion algorithm consists of multiple streams of modality-specific Convolutional Neural Networks (CNNs), which are jointly optimized at multiple feature abstraction levels. Multiple features are extracted at several different convolutional layers from each modality-specific CNN for joint feature fusion, optimization, and classification. Features extracted at different convolutional layers of a modality-specific CNN represent the input at several different levels of abstract representations. We demonstrate that an efficient multimodal classification can be accomplished with a significant reduction in the number of network parameters by exploiting these multi-level abstract representations extracted from all the modality-specific CNNs.  We demonstrate an increase in multimodal person identification performance by utilizing the proposed multi-level feature abstract representations in our multimodal fusion, rather than using only the features from the last layer of each modality-specific CNNs. We show that our deep multi-modal CNNs with multimodal fusion at several different feature level abstraction can significantly outperform the unimodal representation accuracy. We also demonstrate that the joint optimization of all the modality-specific CNNs excels the score and decision level fusions of independently optimized CNNs. 
\end{abstract}


%
\IEEEpeerreviewmaketitle

\section{Introduction}
Feature representation in the biometric systems can be unimodal or multimodal, where the unimodal schemes use a single biometric trait and the multimodal schemes combine the features extracted from multiple biometric feature representations. Benefiting from fusion of features, multimodal biometric models have demonstrated more robustness to noisy data, non-universality and category-based variations~\cite{jaafar2013review,toli2014survey}. However, one of the major challenges in multimodal biometric systems is the fusion algorithm~\cite{nagar2012multibiometric}. The fusion algorithm can be performed at signal, feature, score, rank or decision levels~\cite{connaughton2013fusion,singh2004infrared,nadheen2013feature}, using different schemes such as feature concatenation~\cite{shi2016rule,goswami2016group} and bilinear feature multiplication~\cite{lin2015bilinear,chowdhury2016one,Soleymani2018generalized}.
 
Compared to score, rank, and decision level fusions, feature level fusion results in a better discriminative classifier~\cite{faundez2005data,ross2005feature}, due to preservation of raw information~\cite{shekhar2014joint}. The feature level fusion integrates different features extracted from different modalities into a more abstract and compact feature representation, which can be further used for classification, verification, or identification~\cite{eshwarappa2011multimodal,haghighat2016discriminant}. Recently several authors have exploited feature level fusion for multimodal biometric identification. Among them the serial feature fusion~\cite{liu2001shape}, the parallel feature fusion~\cite{yang2003feature}, the CCA-based feature fusion~\cite{sun2005new}, JSRC~\cite{shekhar2014joint}, SMDL~\cite{bahrampour2016multimodal}, and DCA/MDCA~\cite{haghighat2016discriminant} are the most prominent techniques. 

To integrate features from different modalities, several fusion methods have been considered in the literature~\cite{1221234,lumini2017overview}. One of the major challenges in multimodal fusion is managing the large dimensionality of the fused feature representations, which highlights the importance of the fusion algorithm. The prevalent fusion method in the literature is feature concatenation, which is very inefficient as the feature space dimensionality increases~\cite{nagar2012multibiometric,goswami2016group}. Also it does not explore features at different levels of representation and abstraction. To overcome this shortcoming, the weighted feature fusion and multi-level abstract feature representations of individual modalities are proposed in this paper. Using multi-level feature abstraction, feature descriptors at different feature resolutions and abstractions are considered in the proposed classification algorithms. Our proposed fusion method also enforces the higher level dependencies between the modalities through the joint optimization of modality-specific CNNs and backpropagation algorithm. Similar fusion methods have outperformed the conventional feature fusion methods in applications such as multi-task learning~\cite{ranjan2016hyperface} and gesture recognition~\cite{neverova2016moddrop,neverova2014multi}. 

Convolutional neural networks have been used as classifiers, but they are also efficient tools to extract and represent discriminative features from the raw data at different levels of abstraction. Compared to hand-crafted features, employing CNN as domain feature extractor demonstrated to be more promising when facing different modalities such as face~\cite{kazemi2018attribute,lawrence1997face,iranmanesh2018deep,kazemi2018facial}, iris~\cite{gangwar2016deepirisnet} and fingerprint~\cite{nogueira2016fingerprint,dabouei2018fingerprint}. However, the effects of the fusion at different levels of feature resolution and abstraction and joint optimization of the architecture are not investigated for multimodal biometric identification.  

In this paper, we make the following contributions: (i) rather than fusing the networks at the softmax layer, the optimally compressed feature representations of all modalities are fused at the fully-connected layers without loss of any performance accuracy, but with a significant reduction in the number of network parameters; (ii) instead of spatial fusion at the convolutional layer, modality-dedicated layers are designed to represent the features for later fusion; (iii) the fully data-driven architecture using fused CNNs, is optimized for joint domain-specific feature extraction and representation with the application of person identification; (iv) in the proposed architecture all the CNNs, the joint representation, and the classifier are jointly-optimized. Therefore, a jointly optimized multimodal representation of all the modalities is constructed. In the previous multi-stream biometric state of the art CNN architectures, the modality-dedicated networks are optimized separately, and the classifier is independent of the modality-dedicated networks, finally (v)  multi-level abstract feature fusion for biometric person identification is studied.  

To the best of our knowledge this is the first research effort to utilize multi-stream CNNs for joint multimodal fusion person recognition, which deploys multiple abstraction levels of modalities face, iris, and fingerprint. 
\section{Multi-level feature abstraction and fusion}
Our proposed multimodal architecture consists of multiple CNN-based modality-dedicated networks and a joint representation layer, which are jointly optimized. The modality-dedicated networks are trained to extract the modality specific features at different abstract levels, and the joint representation is trained to explore and enforce dependency between different modalities.

\begin{figure*}
\begin{center}
\includegraphics[width=1\linewidth]{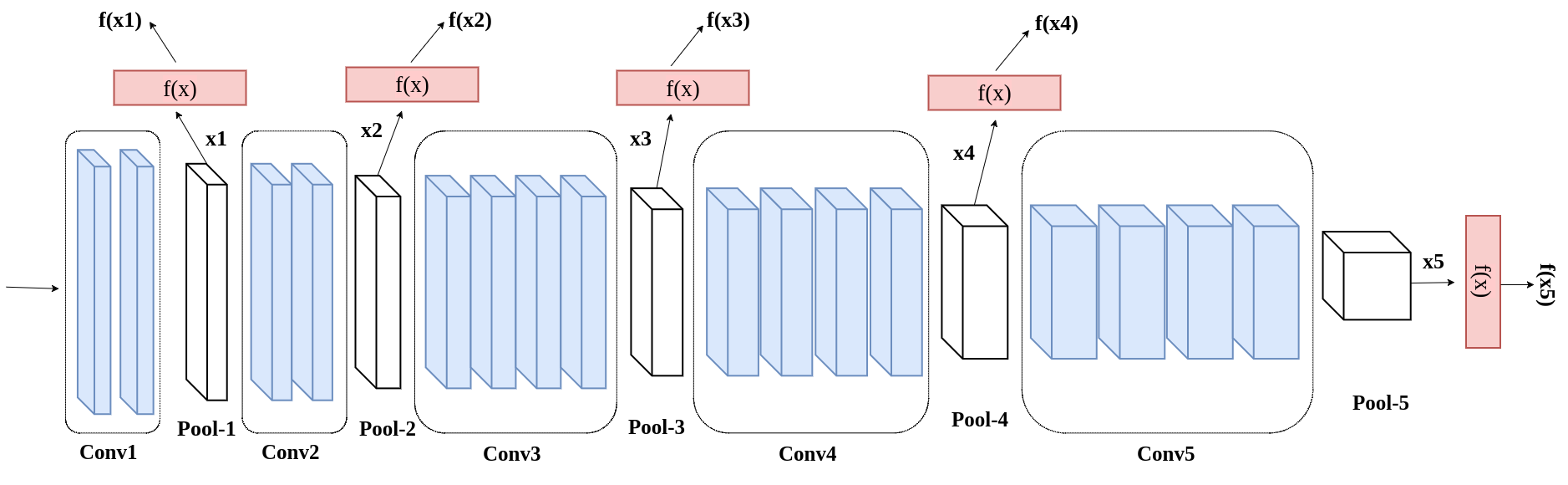}
\end{center}
\caption{Our general scheme for the multi-level feature abstraction from a modality-dedicated network, where $f$ maps the feature space to the one-dimensional feature vector. Therefore, the modality-dedicated network is represented by the modality-dedicated embedding layers.}
\label{fig:1CNN_n}
\end{figure*}

In the CNN architecture, each layer represents different abstract feature representation of the input, where deeper levels provide more complex and abstract features. To benefit from different resolutions and abstractions generated by feature maps at different layers of each modality-dedicated CNN, we propose to utilize the information within the feature maps at different layers in our classification algorithm. One generic example for this approach is presented in Figure~\ref{fig:1CNN_n}, where deep and shallow level feature maps are contributing in the classification algorithm. In this example, function $f$ maps the feature map space to a one-dimensional feature vector. Then, the combination of the vectors extracted from different levels of abstraction are considered for the classification task. 

In this paper, $maxpooling$ followed by a fully-connected layer is considered as the function $f$. Another example for the function $f$ is $globalpooling$, where each feature map is averaged to construct the representative feature vector. These mappings drastically reduce the number of parameters in the model. In this paper, we focus on the first example where the feature space is mapped to a feature vector through $maxpooling$ and one fully-connected layer, as presented in Figure~\ref{fig:1CNNn} and Table~\ref{table:architecture_FC7}. 

In the proposed multi-stream CNN, different levels of abstraction from each modality contribute to the decision making algorithm. Table~\ref{tab:fusion} (b) presents one example for multi-stream multimodal CNN architecture, where each modality is represented in the decision making algorithm by both its deep and shallow feature maps.  
\label{table:dimensionalityreduction}
\section{Modality-dedicated networks}\label{sec:architecture}
Each modality-dedicated CNN consists of the first 16 convolutional layers of VGG19~\cite{simonyan2014very} and a fully-connected dimensionality reduction layer (FC6) of size $1024$. The conventional VGG19 networks are not practical for this application, since the joint optimization of all the modality-dedicated networks and the joint representation is practically impossible, as the result of massive number of parameters that are needed to be trained. Limitations in the number of training samples, along with large feature dimensionality, result in different training phase complexities which require solutions such as Bayesian controlled sampling~\cite{broumand2015discrete}, imposing common structural assumptions on features~\cite{broumand2015length}, and few-shot domain adaptation~\cite{motiian2017few}. 
In the proposed framework, due to the limited number of training samples, it is not applicable to train a vast number of weights in the last layer of the architecture.
Therefore, the number of kernels in the fully connected layer (FC6), compared to the conventional VGG19, is decreased to $1024$.
The details for each modality-dedicated network can be found in Table~\ref{table:architecture2}.
\begin{table}[t]
\begin{center}
\addtolength{\tabcolsep}{-5pt}
\begin{tabular}{l@{\hskip .05in}c@{\hskip .05in}c@{\hskip .05in}c}
\toprule 
network & CNN-Face & CNN-Iris&CNN-Fingerprint\\
\hline
input&$224\times 224\times 3$&$64\times 512\times 3$&$224\times 224\times 3$\\
\toprule 
layer&kernel&kernel&kernel\\
\hline
conv1 (1-2)& $3\times 3 \times 64$ & $3\times 3 \times 64$& $3\times 3 \times 64$\\
\rowcolor{black!10} maxpool1 & 2 $\times 2$ &$ 2 \times 2$&$ 2 \times 2$\\
conv2 (1-2) & $3\times 3 \times 128$ & $3\times 3 \times 128$& $3\times 3 \times 128$\\
\rowcolor{black!10} maxpool2 & $2 \times 2$ & $2 \times 2$& $2 \times 2$\\
conv3 (1-4) & $3\times 3 \times 256$ & $3\times 3 \times 256$& $3\times 3 \times 256$\\
\rowcolor{black!10} maxpool3 & $2 \times 2$ & $2 \times 2$& $2 \times 2$\\
conv4 (1-4) & $3\times 3 \times 512$ & $3\times 3 \times 512$& $3\times 3 \times 512$\\
\rowcolor{black!10} maxpool4 & $2 \times 2 $& $2 \times 2$& $2 \times 2$\\
conv5 (1-4) & $3\times 3 \times 512$ & $3\times 3 \times 512$& $3\times 3 \times 512$\\
\rowcolor{black!10} maxpool5 & $2 \times 2 $& $2 \times 2$& $2 \times 2$\\
FC6 &  $7\times 7\times 1024$ & $2\times 16\times 1024$&$7\times7\times 1024$\\
\bottomrule
\end{tabular}
\end{center}
\caption[Table caption text]{The modality-dedicated CNN architectures. Notation conv3 (1-4) represents all four convolutional layers conv3-1,..., conv3-4, where each of these layers includes $256$ kernels of size $3\times 3$.}   
\label{table:architecture2}
\end{table}
\section{Fusion algorithms}
In this section, we investigate the fusion of multi-stream CNN architectures. The main goal for the fusion layer is to train the multimodal CNNs such that the ultimate joint feature representation outperforms single modality representations. A recognition algorithm using a multimodal architecture requires selecting the discriminative and informative features at different levels from each modality, as well as exploring the dependencies between different modalities. In addition, the joint optimization should discard the redundant single modality features that are not useful in the joint recognition.

\begin{figure*}
\begin{center}
\includegraphics[width=1\linewidth]{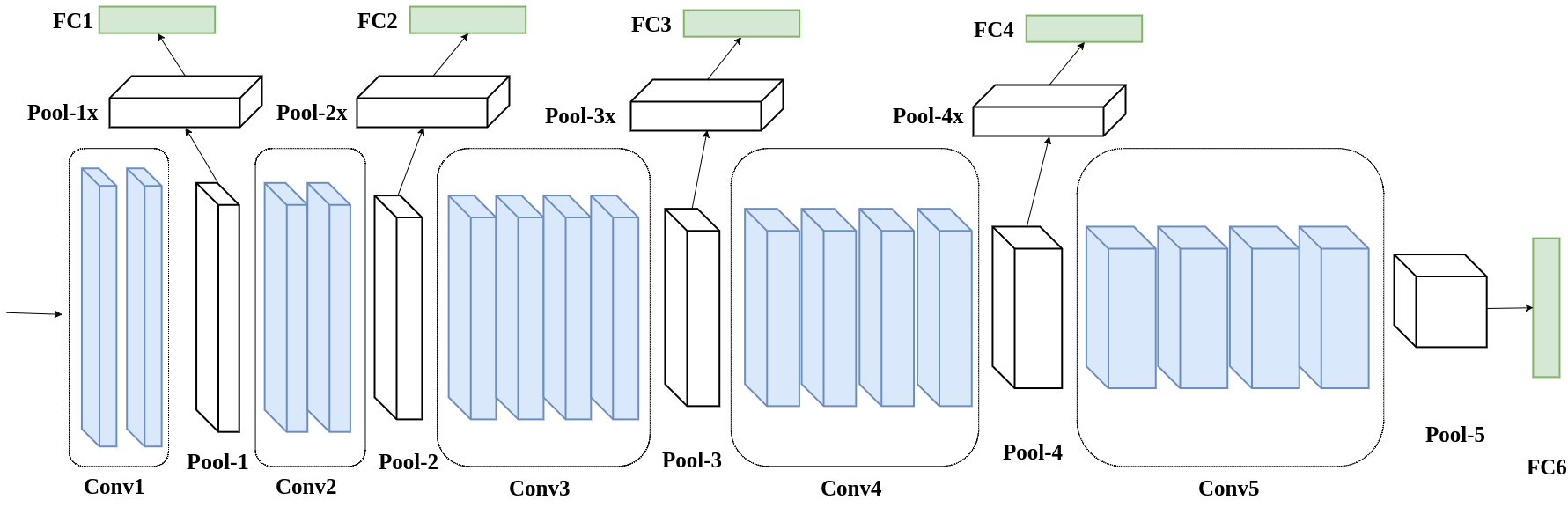}
\end{center}
\caption{Our modality-dedicated network is a CNN which consists of the first 16 layers of a VGG19 network and fully connected layers of size $1024$. In our fusion algorithm, only $pool_{3x}$ and FC3 of size $1024$ are considered for the proposed multi-abstract architecture.}
\label{fig:1CNNn}
\end{figure*}

Fusion can be performed on the feature maps of the CNNs, when the feature maps, corresponding to different modalities have the same spatial dimensions. 
However, in multimodal architectures, the feature level representations can vary in the spatial dimension, due to different inputs' spatial dimensionality. To handle this issue, instead of utilizing feature map layers for fusion, fully-connected layers constructed from different CNN feature maps are used in our architecture for ultimate modality-dedicated feature representation. Prior to the fusion, each modality is represented by either the output of its last fully-connected layer, or from multiple CNN layers representing abstract levels. We call these fully connected layers the {\it modality-dedicated embedding layers}. We demonstrate that the fully-connected representation provides promising results in the case of recognition application. A generic scheme for modality-dedicated embedding layers for a modality-dedicated CNN can be found in Figure~\ref{fig:1CNNn}, where deep and shallow level feature maps are represented by modality-dedicated embedding layers. 

\subsection{Weighted feature fusion} 
In this fusion algorithm, the joint layer is built upon the weighted fusion of the last modality-dedicated embedding layers of the modality-dedicated CNNs. The number of features in this layer is equal to the sum of the number of the output features in the last modality-dedicated embedding layers of the modality-dedicated networks. For instance, for BioCop database~\cite{BIIC} which consists of three modalities, the three modality dedicated embedding layers (FC6 layers in Table~\ref{table:architecture2}) build a layer of size $3072$. Then they are fused together using a fully-connected layer of size $1024$. The output of this fusion layer is fed to the fully-connected classification layer of size $294$, as shown in Table~\ref{tab:fusion}~(a), while no non-linear activation is performed at classification layer. Softmax function is utilized to normalize the outputs of this layer. By training the whole architecture jointly, the first order dependency is enforced between the modalities through backpropagation.


The BIOMDATA database~\cite{crihalmeanu2007protocol}, used in our experiments, consists of four fingerprint and two iris modalities. Considering the nature of this database, we propose to use a bi-level weighted feature fusion. In this fusion algorithm, the four fingerprint modalities are fused together using a fully-connected layer of size $1024$. Similarly, the two iris modalities are fused together using a fully-connected layer of size $1024$. The outputs of these two fully-connected layers are fused using the classification layer of size $219$ as presented in Table~\ref{tab:fusion}~(c).
\subsection{Multi-level feature abstraction and fusion} 
To benefit from the different resolutions generated by different layers of the modality-dedicated CNN, the {\it $pool_3$} layer is down-sampled using $maxpool$ of size $4\times 4$. Then, a fully-connected layer of size $1024$ is considered to represent the shallow level feature maps. This modality-dedicated embedding layer, along with the last layer of the original modality-dedicated network (FC6), are employed as the modality-dedicated embedding layers for the classification task, as presented in Table~\ref{tab:fusion}~(b) and (d). The details for the four architectures considered in this paper are presented in Table~\ref{tab:fusion}.
\begin{table}[t]
\begin{center}
\addtolength{\tabcolsep}{-5pt}
\begin{tabular}{l@{\hskip .05in}c@{\hskip .05in}c@{\hskip .05in}c}
\toprule 
network & CNN-Face & CNN-Iris&CNN-Fingerprint\\
\hline
input(pool3)&$28\times 28\times 256$&$8\times 64\times 256$&$28\times 28\times 256$\\
\toprule 
layer&kernel&kernel&kernel\\
\hline
pool3x& $4\times 4 $ & $4\times 4$& $4\times4$\\
\rowcolor{black!10}FC3 &  $7\times 7\times 1024$ & $2\times 16\times 1024$&$7\times7\times 1024$\\
\bottomrule
\end{tabular}
\end{center}
\caption[Table caption text]{Additional layers added to each modality-dedicated network for multi-level feature abstraction fusion.}   
\label{table:architecture_FC7}
\end{table}

\begin{table}[t]
	\begin{subtable}[h]{0.5\textwidth}
		\centering
		\begin{tabular}{l|c}
		\hline
		Input & FC6$_{m1}$, FC6$_{m2}$, FC6$_{m3}$\\
		\hline
		\rowcolor{black!10}fusion layer & 1024  \\
		\hline
    	classification layer & 294  \\
		\hline
    	\rowcolor{black!10}\multicolumn{2}{c}{softmax}  \\
    	\hline
		\end{tabular}
		\caption{Weighted feature fusion for BioCop database}
		\label{tab:fusion_A}
	\end{subtable}
	\hfill
	\begin{subtable}[h]{0.5\textwidth}
		\centering
		\begin{tabular}{l|c}
		\hline
		Input & FC3$_{m1}$, FC3$_{m2}$, FC3$_{m3}$,\\

		& FC6$_{m1}$, FC6$_{m2}$, FC6$_{m3}$\\		
		\hline
		\rowcolor{black!10}fusion layer & 1024  \\
		\hline
    	classification layer & 294  \\
		\hline
    	\rowcolor{black!10}\multicolumn{2}{c}{softmax}  \\
    	\hline
		\end{tabular}
		\caption{Multi-abstract fusion for BioCop database}
		\label{tab:fusion_B}
	\end{subtable}
	\hfill
	\begin{subtable}[h]{0.5\textwidth}
		\centering
		\begin{tabular}{l|c|c}
		\hline
		Input & FC6$_{n1}$, FC6$_{n2}$, & FC6$_{n5}$, FC6$_{n6}$\\
        & FC6$_{n3}$, FC6$_{n4}$& \\		
		\hline
		\rowcolor{black!10}fusion layers & 1024 & 1024\\
		\hline
    	classification layer & \multicolumn{2}{c}{219}  \\
		\hline
    	\rowcolor{black!10}\multicolumn{3}{c}{softmax}  \\
    	\hline
		\end{tabular}
		\caption{Bi-level weighted feature fusion for BIOMDATA database}
		\label{tab:fusion_C}
	\end{subtable}
	
	\hfill
	\begin{subtable}[h]{0.5\textwidth}
		\centering
		\begin{tabular}{l|c|c}
		\hline
		Input & FC3$_{n1}$, FC3$_{n2}$, FC3$_{n3}$, &FC3$_{n5}$, FC3$_{n6}$,\\
		      & FC3$_{n4}$, FC3$_{n1}$, FC6$_{n2}$, &FC6$_{n5}$, FC6$_{n6}$\\
		      & FC6$_{n2}$, FC6$_{n3}$, FC6$_{n4}$&\\
		\hline
		\rowcolor{black!10}fusion layers & 1024 & 1024\\
		\hline
    	classification layer& \multicolumn{2}{c}{219}  \\
		\hline
    	\rowcolor{black!10}\multicolumn{3}{c}{softmax}  \\
    	\hline
		\end{tabular}
		\caption{Bi-level multi-abstract fusion for BIOMDATA database}
		\label{tab:fusion_D}
	\end{subtable}
	\caption{Joint representation architectures and modality-dedicated embedding layers for BioCop (a and b) and BIOMDATA (c and d) databases. $m_1$, $m_2$, and $m_3$ represent three BioCop database modalities. $n_1$, $n_2$, $n_3$, and $n_4$ represent four fingerprint modalities, and $n_5$ and $n_6$ represent two iris modalities for BIOMDATA database.}
	\label{tab:fusion}
\end{table}
\label{table:dimensionalityreduction}
\section{Experimental Setup}\label{sec:Setup}
To evaluate the performance of the proposed fusion multimodal architectures, two challenging multimodal biometric databases are considered:
\subsection{Datasets}\label{Dataset}
{\bf {BioCop multimodal database:}} The proposed algorithm is evaluated on BioCop database~\cite{BIIC}. This database is one of the few databases that allows disjoint training and testing of multimodal fusion at feature level. The BioCop dataset is collected under four disjoint years; 2008, 2009, 2012 and 2013. Each label consists of different biometric modalities for each subject; face, iris, fingerprint, palm print, hand geometry, voice and soft biometrics. To evaluate the performance of the proposed architectures, the following three biometric modalities are considered in this experiment: face, iris, and right index fingerprint. 

Under each label, the biometrics are acquired during either one or two separate sessions. The 2012 and 2013 databases contain 1,200 and 1,077 subjects, respectively. To make the training-test splits mutually exclusive, the 294 subjects that are common in labels 2012 and 2013 are considered. The proposed algorithms are trained on 294 mutual subjects in year 2013 dataset, and are tested on the same subjects in year 2012 dataset. It worths mentioning that although the databases are labeled as 2012 and 2013, the date of data acquisition for common subjects in the datasets can vary between one to three years, which also adds the advantage of investigating the age-progression effect. We have also considered the left and right irises as a single class, which results in heterogeneous classes for the iris modality.

In both 2012 and 2013 databases, for each individual, the number of samples per modality may vary. Therefore, in each database, for each individual, $250$ triplet of modalities are randomly chosen. Each triplet includes preprocessed face, iris, and right index fingerprint images. The number of image triplets in both training and test sets is the same, and equal to $73,500$.

{\bf BIOMDATA multimodal database}: BIOMDATA database~\cite{crihalmeanu2007protocol} is a challenging database, since many of the image samples are damaged with blur, occlusion, sensor noise and shadows~\cite{haghighat2016discriminant}. This database is a collection of biometric modalities: iris, face, voice, fingerprint, hand geometry, and palm print, from subjects of different ethnicity, gender, and age. Due to privacy issues, face data is not available in combination with other modalities. To evaluate the performance of the proposed architectures, the following six biometric modalities are considered in this experiment: left and right iris, and thumb and index fingerprints from both hands. 

The experiments are conducted on 219 subjects that have samples in all six modalities. For each modality, four randomly chosen samples are considered for the training phase and the remaining samples are considered for the test phase. For any modality in which the number of the samples is less than five, one sample is considered for the test phase and the remaining samples are considered for training. The summary of the considered databases is presented in Table~\ref{table:datasetsBIOMDATA}.
\begin{table}[t]
\begin{center}
\begin{tabular}{c| l c c}
\toprule 
\multicolumn{1}{c}{} & \multicolumn{1}{c}{} & \multicolumn{1}{c}{Train set} & \multicolumn{1}{c}{Test set} \\ \hline
\hline
\multirow{3}{*}{\rotatebox[origin=c]{90}{BioCop}}&Face&6833&6960\\
& Iris&36636&39725\\
&Fingerprint&1822&991\\
\hline
\multirow{6}{*}{\rotatebox[origin=c]{90}{BIOMDATA}}&Left iris &874&584\\
&Right iris&871&581\\
&Left thumb&875&644\\
&Left index&872&632\\
&Right thumb&871&647\\
&Right Index&870&624\\
\bottomrule
\end{tabular}
\end{center}
\caption[Table caption text]{The size of the training and test sets for each modality in BioCop and BIOMDATA databases.}
\label{table:datasetsBIOMDATA}
\end{table}
For both the training and test sets, for each individual, $250$ set of samples are randomly chosen, where each set includes normalized left and right irises, and enhanced left index, right index, left thumb and right thumb fingerprint images. The number of samples in both training and test sets is the same, and equal to $54,750$.
\subsection{Preprocessing}
\label{preprocessing}
For the face modality, the frontal images are considered. The face images are cropped, aligned to a template~\cite{dlib09}, and resized to $224\times 224$ images. Fingerprint images are enhanced using the method
described in~\cite{chikkerur2004systematic}. The core point is detected from the enhanced images~\cite{jain2000filterbank}. Finally, the $224 \times 224$ region centered by the core point is cropped.
  
Iris images are segmented and normalized using OSIRIS~\cite{krichen2008osiris}. Although OSIRIS software does not mask eyelids and eyelashes, the segmented images do not contain much occlusion due to eyelids~\cite{hollingsworth2009best}. OSIRIS algorithm finds the iris inner and outer contours. This two contours are used to transform the iris area into a $64\times 512$ strip.
\section{Joint Optimization of Networks}
In this section, the training of the multimodal CNN architecture is discussed. Here, we explain the implementation of each modality-dedicated network, the joint fusion representation layer, and the concurrent optimization of the multimodal CNNs and the fusion layer. The fusion layer can be either the weighted fusion layer or multi-abstract fusion layer.
\subsection{Modality-dedicated networks}
Initially, the modality-dedicated CNNs are trained independently and each CNN is optimized for a single modality. As explained in section~\ref{sec:architecture}, each of these CNN networks consists of the first 16 convolutional layers of VGG19 network with an added fully-connected feature reduction FC6 layer of size $1024$. The extra layer is dedicated to make the feature level fusion tractable. For each modality, the conventional VGG19 network is trained as explained below. For all the modalities, the networks are initialized by VGG19 pre-trained on Imagenet~\cite{deng2009imagenet}.

{\bf CNN-Face:} To optimize the weights for extracting the face features, the pre-trained network is fine-tuned on the CASIA-Webface~\cite{yi2014learning} and the BioCop~\cite{BIIC} face 2013. The network is then trained on $294$ subjects in the dataset 2013 that are also present in the 2012 dataset. Finally, previously trained weights for the first 16 layers of the network, along with FC6 layer of size $1024$ and the softmax layer, are fine-tuned on the $294$ subjects in 2013 dataset.

The face image inputs are $224\times 224$ RGB images. The preprocessing algorithm consists of the channel-wise mean subtraction on RGB values, where channel means are calculated on the whole training set. The training algorithm is deployed by minimizing the softmax cross-entropy loss using mini-batch stochastic gradient descent with momentum. The training is regularized by weight decay and $50\%$ dropout for the fully-connected layers except for the last layer. The batch size, momentum and L$_2$ penalty multiplier are set to 32, 0.9, and $0.0005$, respectively. The initial learning rate is set to $0.1$. The learning rate is decreased exponentially by a factor of $0.1$ for every $2$ epochs of training. In this network, batch normalization~\cite{ioffe2015batch} is applied. The moving average decay is set to $0.99$.

{\bf CNN-Iris:} Similar to the face network, the training is performed over the Imagenet pre-trained VGG19. To specify the kernels to extract the iris-specific features, the pre-trained network is finetuned on the CASIA-Iris-Thousand~\cite{CASIA-IRIS} and Notre Dame-IRIS 04-05~\cite{bowyer2016nd}.

For the BioCop~\cite{BIIC} database, the network is then tuned on BioCop iris 2013. The network is then fine-tuned on $294$ subjects in the dataset 2013 which are also present in the 2012 dataset. After dropping the last two layers of VGG19 and adding FC6 and the softmax layers, the network is once again trained on these 294 subjects in 2013 dataset. The iris image inputs are $64\times 512$ grayscale images. The optimization parameters are the same as face architecture. In this network, batch normalization is also applied. The moving average decay is set to $0.9$. The learning rate decrease exponentially by a factor of $0.1$ every $5$ epochs.

For the BIOMDATA database~\cite{crihalmeanu2007protocol}, for each of the two modalities, the pre-tuned network is tuned on all subjects that have samples in that modality, and then, on 219 subjects that have samples in all six modalities. Since the number of samples in this dataset is much smaller than the number of samples in BioCop database, the learning rate decay is set to $0.99$. Similar to CNN-Face, each tuned networks is fine-tuned after dropping the fully connected layers and adding FC6.

{\bf CNN-Fingerprint:} Fingerprint networks are initiated with Imagenet pre-trained VGG19 weights. For BioCop database, it is then trained on BioCop 2013 fingerprint dataset. Then, the network is fine-tuned on $294$ subjects in the dataset 2013 which are also present in the 2012 dataset. For the BIOMDATA database, for each of the four modalities, the pre-tuned network is tuned on all subjects that have samples in that modality, and then, on 219 subjects that have samples in all six modalities. The inputs are $224\times 224$ grayscale images. The optimization parameters are the same as CNN-Face architecture. The learning rate decreases exponentially by a factor of $0.1$ every $10$ epochs. 
\subsection{Joint optimization of networks}\label{trainjointrepresentation}
Initially, to train the joint representation, the modality-dedicated networks' weights are frozen, and the joint representation layer is optimized greedily upon the extracted features from the modality-dedicated networks. The optimization parameters are the same as the CNN-Fingerprint network. Finally all the networks are jointly optimized. Here, the batch size is further reduced, 
and the initial learning rate is reduced to the the smallest final learning rate among modality-dedicated networks. In all the mentioned steps, Rectified Linear Unit (ReLU) activation function is utilized for all the layers except the classification layer. 
\subsection{Hyperparameter optimization}
The hyperparameters in our experiments are : $\lambda$ the regularization parameter, $\alpha_0$ initial learning rate, $n$ number of epochs per decay for the learning rate, $d$ moving average decay, and the $m$ as the momentum. For each optimization, the five-fold cross-validation method is considered to estimate the best hyperparameters.
\begin{figure*}[t]
\begin{subfigure}{.5\textwidth}
\begin{center}
\includegraphics[width=1\linewidth]{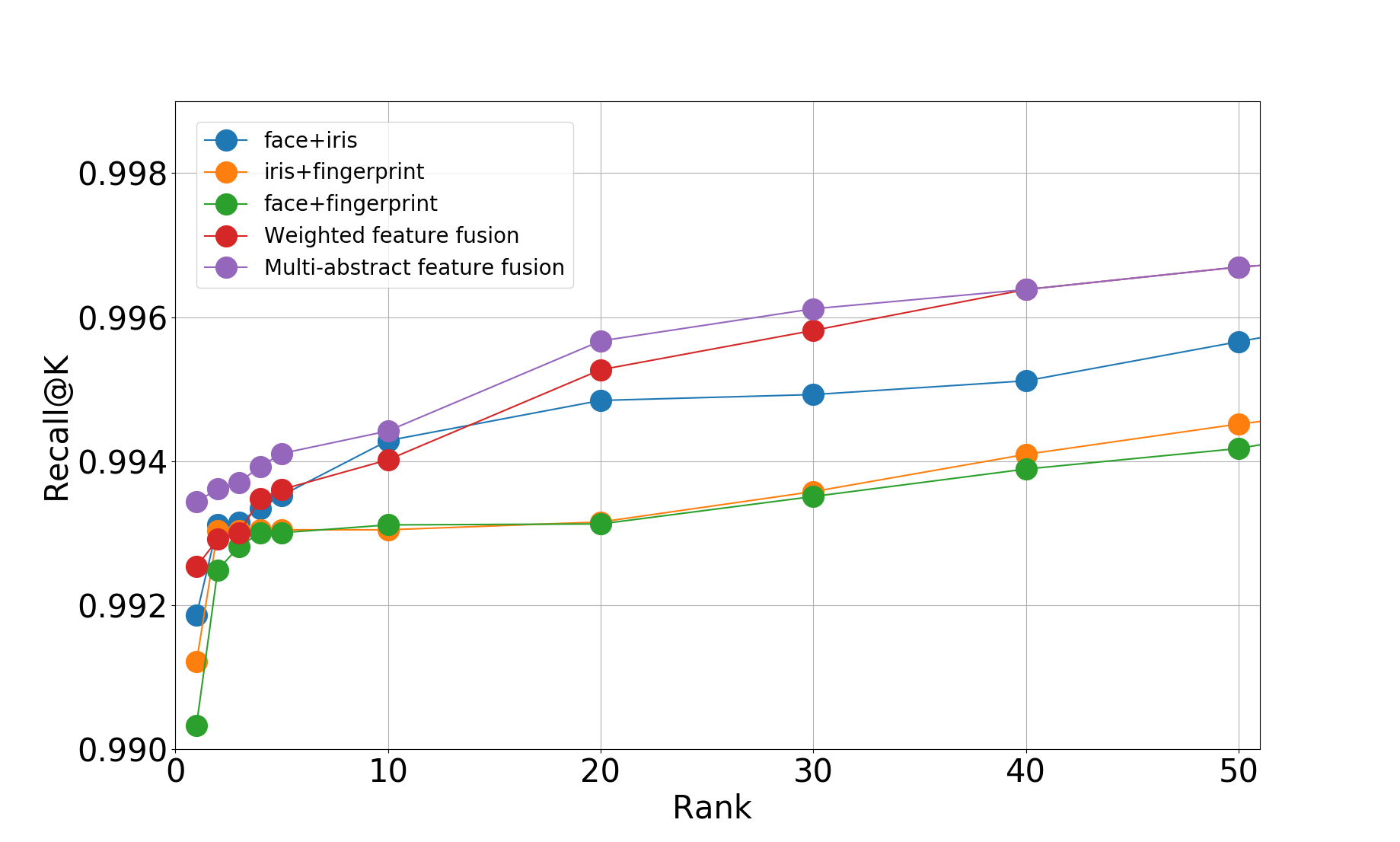}
\end{center}
\caption{}
\label{fig:fusion1}
\end{subfigure}
\begin{subfigure}{.5\textwidth}
\begin{center}
\includegraphics[width=1\linewidth]{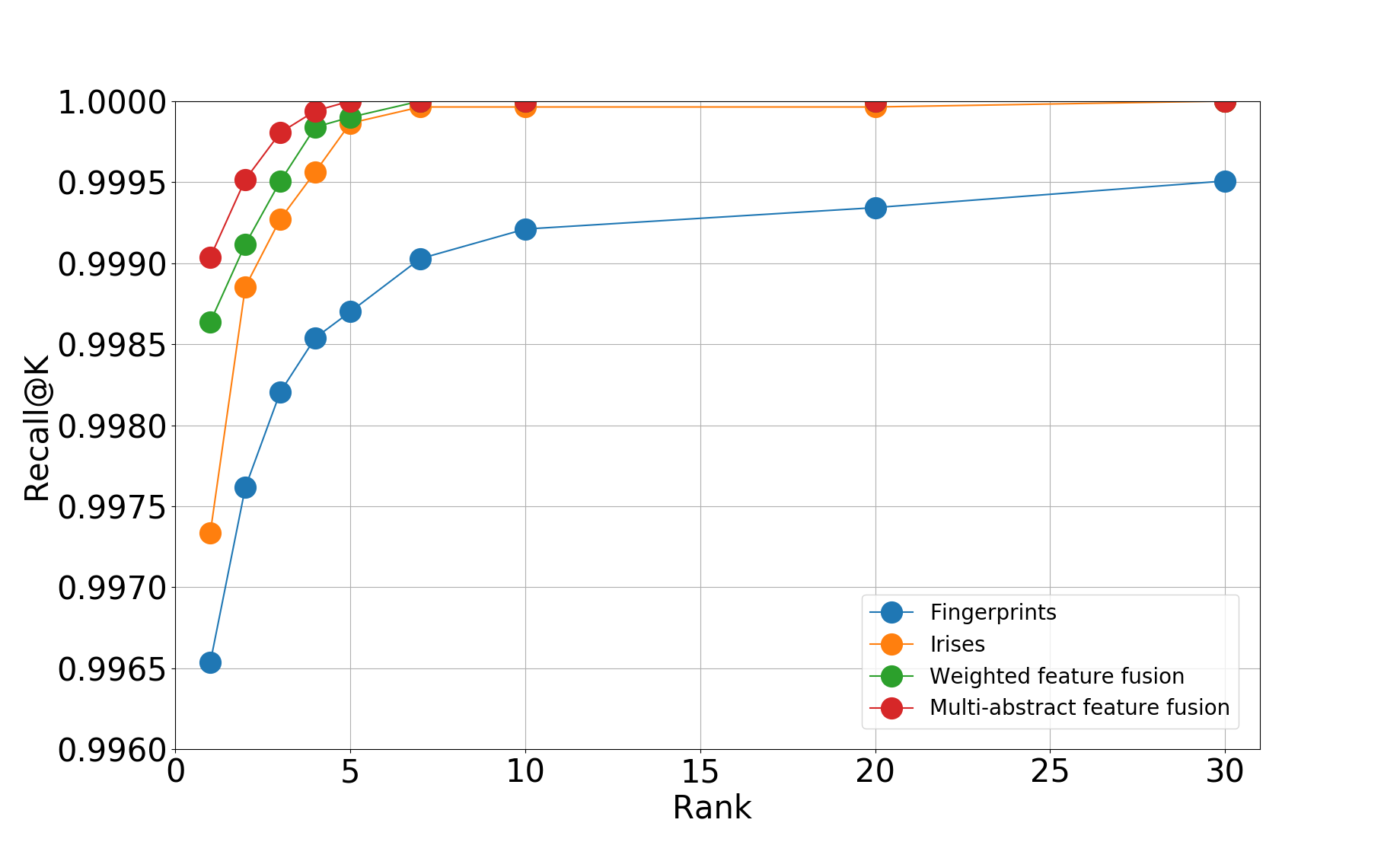}
\end{center}
\caption{}
\label{fig:fusion2}
\end{subfigure}
\caption{CMC curves for (a) BioCop, and (b) BIOMDATA databases.}
\label{fig:fusion2vs1}
\end{figure*}

\section{Experiments and disscussions}\label{sec:Experiments}

\subsection{Evaluation metrics}\label{sec:evaluation}
The performance of different experiments are reported and compared using two classification metrics: classification accuracy and \emph{Recall@K}. The classification accuracy is the fraction of correctly classified samples regarding their classes. The \emph{Recall@K} metric is the probability that a subject class is correctly classified at least at rank-k, while the candidate classes are sorted by their similarity score to the query samples.~The calculation of \emph{Recall@K} is done per class, and is averaged over all available classes. 

The reported values are the average values for five randomly generated training and test sets. As explained in section~\ref{sec:Setup}.~{\it A}, training and test sets consist of $73,500$ triplet images for BioCop database and $54,750$ sets of six images for BIOMDATA database. 

\subsection{Data augmentation}
For both databases, data augmentation is performed on the fingerprint images. 20 augmented samples of each fingerprint image is generated by translating the core point both vertically and horizontally using Gaussian distribution. For each fingerprint image, ten augmented images are generated using Gaussian distribution with parameters $\mu=0$ and $\sigma=2.5$. The remaining ten augmented images are generated with $\mu=0$ and $\sigma=5$ being considered. In Table~\ref{table:results_NBIS} studies the effect of data augmentation on the rank-one recognition rate for the modality-dedicated CNN-Fingerprint for BioCop database. This table also includes the recognition rate for NBIS software~\cite{ko2007users}.
\begin{table}[t]
\begin{center}
\begin{tabular}{l c c c}
\toprule 
 & NBIS &CNN w/o &CNN\\
\hline
Right index&95.67&96.08&97.28\\
\hline
\end{tabular}
\end{center}
\caption[Table caption text]{Rank-one recognition rate for BioCop database utilizing NBIS software, CNN without data augmentation and CNN with data augmentation.}
\label{table:results_NBIS}
\end{table}
\subsection{Results}\label{Results}
To compare the results for the proposed algorithms with the state-of-the-art algorithms, Gabor features in five scales and eight orientations are extracted from all modalities. For the face images, $31,360$ features are extracted from $224\times 224$ aligned images. While, for the iris images, $36,630$ features are extracted from $64 \times 512$ segmented and normalized image. In the case of fingerprint images, $31,360$ features are extracted from the enhanced $224\times 224$ images, as described in Section~\ref{preprocessing}, around the core point. These features are used for all the state-of-the-art algorithms except CNN-Sum, CNN-Major, and two proposed algorithms.

Table~\ref{table:results_single} presents single modality rank-one recognition rate for both the databases. In this table the Gabor features are used for SVM and KNN algorithms. The performance of the proposed weighted feature level fusion and multi-abstract fusion algorithms are compared with that of several state-of-the-art feature, score and decision level fusion algorithms in Tables~\ref{table:results_Biocop} and~\ref{table:results_BIOMDATA}. SVM-Sum and CNN-Sum use the probability outputs for the test sample of each modality, added together to produce the final score vector. SVM-Major and CNN-Major chose the maximum number of modalities taken to be from the correct class. 
\begin{table}[t]
\begin{center}
\begin{tabular}{c| l c c c}
\toprule 
\multicolumn{1}{c}{} & \multicolumn{1}{c}{} & \multicolumn{1}{c}{KNN} & \multicolumn{1}{c}{SVM} &\multicolumn{1}{c}{CNN}\\ \hline
\hline
\multirow{3}{*}{\rotatebox[origin=c]{90}{BioCop}}&Face&89.68&88.76&98.14\\
&Iris&70.52&79.26&99.05\\
&Right index&91.22&90.61&97.28\\
\hline
\multirow{6}{*}{\rotatebox[origin=c]{90}{BIOMDATA}}&Left iris &66.61&71.92&99.35\\
&Right iris&64.89&71.08&98.95 \\
&Left thumb&61.23&63.96 &80.15 \\
&Left index&82.91&84.70 &93.43 \\
&Right thumb&62.11&63.52&82.63\\
&Right Index&82.05&84.46&93.12\\
\bottomrule
\end{tabular}
\end{center}
\caption[Table caption text]{Rank-one recognition rate for single modalities.}
\label{table:results_single}
\end{table}

The feature level fusion techniques include the serial feature fusion~\cite{liu2001shape}, the parallel feature fusion~\cite{yang2003feature}, the CCA-based feature fusion~\cite{sun2005new}, JSRC~\cite{shekhar2014joint}, SMDL~\cite{bahrampour2016multimodal} and DCA/MDCA~\cite{haghighat2016discriminant} algorithms. Note that in case of more than two modalities, the parallel feature fusion method cannot be applied. Tables~\ref{table:results_Biocop} and~\ref{table:results_BIOMDATA} present the results for BioCop and BIOMDATA databases, respectively. As presented in this table, both fusion algorithms outperform single-modality and two-modality architectures for BioCop database. Similarly, both the fusion algorithms outperform single-modality, two irises, and four fingerprints architectures. The proposed algorithms excel the score-level and the decision-level fusion algorithms for the independently-optimized CNNs as well. 

Figure~\ref{fig:fusion2vs1} presents Cumulative Match Curves (CMCs) for both databases. Figure~\ref{fig:fusion2vs1}~(a) compares \emph{Recall@K} for two and three modality weighted feature fusion algorithms with the three modality multi-abstract feature fusion algorithm. Similarly, Figure~\ref{fig:fusion2vs1}~(b) compares three weighted feature fusion scenarios (two irises, four fingerprints, and all six modalities) with the six modality multi-abstract feature fusion algorithm. For both the studied databases, the multi-abstract fusion algorithm excels the weighted fusion algorithm both in terms of rank-one recognition rate and CMC curve.
\begin{table}[t]
\begin{center}
\addtolength{\tabcolsep}{-0pt}
\begin{tabular}{lccccc}
\toprule 
Modality & $\{$1,2$\}$&$\{$1,3$\}$&$\{$2,3$\}$&$\{$1,2,3$\}$\\
\hline
SVM-Major & 79.22 & 89.27&80.47 &90.32\\
Serial + PCA + KNN&71.12& 86.28&75.69&76.18\\
Serial + LDA + KNN&80.12&91.28&79.69&82.18 \\
Parallel + PCA + KNN&74.69&88.12&77.58&-\\
Parallel + LDA + KNN&82.53&93.21&82.56&- \\
CCA + PCA + KNN&87.21&95.27&86.44&95.33\\
CCA + LDA + KNN&89.12&95.41&86.11&95.58\\
DCA/MDCA + KNN&83.02&96.36&83.44&86.49\\
\hline
CNN-Sum &99.10 & 98.85& 98.92&99.14\\
CNN-Major &98.51&97.70&98.31&99.03\\
Weighted feature fusion& 99.18&99.03&99.12&99.25\\
Multi-abstract fusion&99.31&99.16&99.20&99.34\\
\bottomrule
\end{tabular}
\end{center}
\caption[Table caption text]{Rank-one accuracy evaluation on BioCop database, for different fusion settings. 1, 2, and 3 represent face, iris, and fingerprint, respectively.}   
\label{table:results_Biocop}
\end{table}

\begin{table}[t]
\begin{center}
\addtolength{\tabcolsep}{-0pt}
\begin{tabular}{lccc}
\toprule 
Modality & 2 irises & 4 fingerprints& 6 modalities\\
\hline
SVM-Major &78.12  &88.34 &93.31 \\
SVM-Sum   &81.23      &94.13     &96.85     \\
Serial + PCA+ KNN &72.31&90.71&89.11 \\
Serial + LDA+ KNN&79.82&92.62&92.81 \\
Parallel + PCA+ KNN&76.45&-&-\\
Parallel + LDA+ KNN&83.17&-&-\\
CCA + PCA + KNN&88.47&94.72&94.81\\
CCA + LDA + KNN&90.96&94.13&95.12 \\
JSRC&78.20&97.60&98.60\\
SMDL&83.77&97.56&99.10\\
DCA/MDCA + KNN&83.77&98.1&99.60\\
\hline
CNN-Sum&99.54&99.46&99.82\\
CNN-Major&99.31&99.42&99.48\\
Weighted feature fusion&99.73&99.65&99.86\\
Multi-abstract fusion&99.81&99.72&99.91\\
\bottomrule
\end{tabular}
\end{center}
\caption[Table caption text]{Rank-one accuracy evaluation on BIOMDATA database, for different fusion settings.}       
\label{table:results_BIOMDATA}
\end{table}
\section{Conclusion}
In this paper, we proposed a joint CNN architecture with feature level fusion for multimodal recognition using multiple modalities of face, iris, and fingerprint. We proposed a multi-abstract network to handle the spatial mismatch problem and yet having no loss in performance with significant reduction in network parameters. We demonstrated that the proposed multi-stream CNNs with multimodal fusion at different feature level abstraction and jointly optimization of modality-dedicated networks, joint representation, and classifier, significantly improve unimodal representation accuracy by incorporating the captured multiplicative interactions of the low-dimensional modality-dedicated feature representations. Two fusion methods at the fully-connected layer are studied, and it is concluded that the multi-abstract fusion outperforms the weighted feature fusion algorithm.  
\begin{center}
ACKNOWLEDGEMENT
\end{center}

This work is based upon a work supported by the Center for Identification Technology Research and the National Science Foundation under Grant $\#1650474$.

\bibliographystyle{IEEEtran}
\bibliography{bib}
\end{document}